\documentclass{article}
\usepackage{spconf,amsmath,graphicx}

\usepackage{newtxtext}
\usepackage{newtxmath}

\usepackage{url}
\usepackage{stfloats}
\usepackage[table]{xcolor}

\usepackage{subcaption}
\usepackage{booktabs}

\usepackage{xcolor}

\definecolor{commentcolor}{RGB}{0,0,0}

\def\Rb{\mathbb{R}}

\def\Xc{{\mathcal{X}}}

\def\Zc{{\mathcal{Z}}}

\def\x{{\mathbf{x}}}
\def\y{{\mathbf{y}}}
\def\z{{\mathbf{z}}}

\def\Lc{{\mathcal{L}}}

\def\Nc{{\mathcal{N}}}

\def\Rb{\mathbb{R}}
\def\Eb{\mathbb{E}}
\def\Uc{\mathcal{U}}

\title{Unsupervised Domain Adaptation for Enhanced Radiometer Image Precipitation Estimation using Conditional Flow Matching}
\name{Author(s) Name(s)}
\address{Author Affiliation(s)}

\name{
	\begin{tabular}[t]{c}
		Victor Enescu$^{1,2}$,
		Assaad Zeghina$^{1,2}$,
		Matthieu Meignin$^{1,2}$,
		Nicolas Viltard$^{1}$,
		Cécile Mallet$^{1,2}$
	\end{tabular}
}

\address{
	\begin{tabular}[t]{c}
		$^{1}$ LATMOS/IPSL, UVSQ Université Paris-Saclay, Sorbonne Université, CNRS
		Guyancourt, France\\[0.3em]
		$^{2}$ ARCHES Team, Inria
		Paris, France
	\end{tabular}
}

\begin{document}
	%\ninept
	%
	\maketitle

	\begin{abstract}
		Deep generative networks have recently achieved unprecedented performance in precise image and video editing using sophisticated textual prompts. However, the effectiveness of such models heavily depends on access to very large supervised and annotated image datasets, which can be very difficult to obtain. This is particularly true for satellite instruments, which very rarely overlap with labelled data, and suffer from domain shifts in the rare occasions they do. In this paper, we investigate the potential of flow matching models for unsupervised domain adaptation of satellite radiometer images. Our main contribution is a novel unsupervised method that achieves precise domain alignment by leveraging parts of the deterministic ordinary differential equations in flow matching models, conditioned on different satellite instruments. A key strength of our approach is its ability to preserve essential information while adapting across any domains since the perturbations are in theory bijective. Extensive experiments conducted on the GPM-Core constellation show the benefit of our conditional domain adaptation, particularly in improving rain precipitation estimation from radiometer imagery.
	\end{abstract}

	\section{Introduction}
	Precipitation estimation from satellite imagery holds tremendous potential for both scientific and societal applications, as vast amounts of data are continuously acquired worldwide. Moreover, unlike ground-based radar observations — which are often sparse or entirely absent in many developing countries — satellite constellations provide near–real-time, global-scale coverage of the Earth’s surface. This capability is crucial not only for improving  water resource management, but also for strengthening our ability to predict extreme hydrological events, such as major floods. This is particularly true for the Global Precipitation Measurement (GPM) mission~\cite{gpm_unified_precipitation_estimation_from_space}, an international collaboration that provides unprecedented spatiotemporal coverage of global precipitation through a constellation of twelve satellites, primarily equipped with passive microwave radiometers. Within this constellation, the GPM Core Observatory (GPM-Core) plays a vital role, as it uniquely carries a dual-frequency precipitation radar (DPR) capable of directly estimating raining rates (RR), alongside the GPM Microwave Imager (GMI), which measures brightness temperatures (TB). This contrasts with the other satellites in the constellation, which are equipped only with passive microwave radiometers, and can measure only brightness temperatures.  \\
	\indent This unique feature of the GPM-Core satellite opens up promising opportunities for deep learning approaches that exploit the combined DPR and GMI measurements to enhance precipitation estimation. This is particularly true for the methods~\cite{drain_extended, gprof_nn}, employing deep neural networks to predict rain rates using the GPM-Core satellite brightness temperatures, and which have  shown the capability to outperform classical approaches like GPROF~\cite{evolution_of_gprof_algo}. Among these methods, the strength of DRAIN~\cite{drain_extended} lies in its convolutional architecture, which exploits the spatial structure of 2D images rather than relying on independent per-pixel estimates as in GPROF — and consequently it often achieves superior performance. This approach is particularly compelling given that the swath of the DPR measurements on the GPM-Core satellite is nearly four times narrower than that of the GMI, enabling rain rate estimation over much larger spatial domains. However, despite those very impressive performances, the DRAIN algorithm does not generalize to other satellites from the constellation and achieves much poorer results when using the corresponding brightness temperatures. This is generally explained by the different orbit characteristics of the employed instruments such as channels, angle of view, or pixel resolutions. Although GPM-Core shares overlapping orbits with those other satellites, training a supervised domain-transfer algorithm is challenging, as these very scarce overlaps occur almost exclusively near $60^\circ$ latitude north and south, making the model prone to geographical biases. Additionally, varying temporal offsets of up to 15 minutes exist between acquisitions, making those observations suitable only as pseudo ground truths. \\
	\indent To tackle this challenge of non-colocated data in satellite imagery, more general strategies often turn to unsupervised domain adaptation (UDA), which has the key advantage of not requiring corresponding ground truth images. These approaches aim to transform the content of an image to match a target distribution by leveraging deep generative neural networks. Traditionally, such methods have heavily relied on Generative Adversarial Networks (GANs)~\cite{gan, i2i_translation_using_cond_adv_nets}, and in particular CycleGAN~\cite{cycle_gan}, which allows for indirect supervision through a cycle-consistency reconstruction loss. This is the case for the methods in~\cite{cycle_gan_drain_f18_gpm}, which use CycleGANs along with the DRAIN~\cite{drain_extended} to predict raining rates from the DMSP-F18 satellite of the Defense Meteorological Satellite Program. More precisely the images obtained with its SSMI/S microwave imager are modified so as to very closely match the distribution of those from GMI. Then, the DRAIN algorithm is run on those adapted images to obtain raining rates. By doing so, very promising results are obtained in~\cite{cycle_gan_drain_f18_gpm}, and ultimately, the only drawback of the method is the notoriously unstable training of GANs~\cite{gan} — which makes it very challenging for this method to be extended to other satellites from the constantly evolving GPM constellation. \\
	\indent In order to address these aforementioned issues, we present a novel unsupervised domain adaptation method, based on the  modern Flow Matching~\cite{flow_matching, stochastic_interpolant, cfm_tong} paradigm so as to adapt brightness temperatures from the F18 satellite to those of the GPM instrument. The main advantage of this approach resides in the bijective nature of flow based models, which possess the remarkable ability of \emph{"ideally"} noising and denoising an image while preserving the most salient physical and structural information of the original image. Extensive experiments showcase the benefits of our approach, matching \textcolor{commentcolor}{or outperforming} other state-of-the-art approaches while being much easier to train.
	
	\section{Related Work}
	Deep generative models have widely been explored for unsupervised domain adaptation. Among them, GAN-based approaches~\cite{gan}, particularly CycleGAN~\cite{cycle_gan} and related image-to-image translation methods~\cite{i2i_translation_using_cond_adv_nets}, have historically achieved strong results. However their main drawback is the training instability and lack of diversity frequently observed with adversarial training objectives~\cite{gan}. More recently, with the emergence of diffusion models~\cite{ddpm, ddim}, novel interesting approaches have emerged as very competitive alternatives. Those include approaches like Flow Matching~\cite{flow_matching, stochastic_interpolant, cfm_tong}, which are inspired by continuous normalizing flows~\cite{node_nf} and retain the bijective properties inherited from normalizing flows~\cite{nice}, while enabling simulation-free training and significantly faster sampling. Within this line of work, some very recent approaches go beyond the standard Gaussian assumption by allowing arbitrary source and target distributions~\cite{cfm_tong, schrodinger_bridge_diffusion_bortoli, schrodinger_bridge_forward_and_backward_sde}, thereby enabling the learning of a pushforward mapping directly between different image distributions. Recent related works also investigate unsupervised domain alignment through structured latent decompositions with the flow matching objective~\cite{serpent_flow}. Alternatively, some works exploit the bijective nature of neural networks for unsupervised domain adaptation, most notably~\cite{align_flow} employs normalizing flows with adversarial losses, and~\cite{ddib} reverses ordinary differential equations (ODEs) for diffusion models. The work in~\cite{ddib} is closely related to our approach but differs in three key aspects: (i) it trains a separate model for each modality rather than relying on conditional generative models; (ii) it integrates the entire ODE solver to map images fully to Gaussian noise, whereas we integrate only a portion of the trajectory to optimally add noise while preserving image content, and this also differs from stochastic partial noising strategies such as~\cite{sdedit}; and (iii) it adopts diffusion instead of the more recent flow-matching paradigm, which is known to yield straighter trajectories, faster integration, and improved image quality~\cite{flow_matching}.
	Building on these advantages, we explore conditional flow-matching–based generative models as an alternative for unsupervised domain adaptation, motivated by their stable training, invertible bijective mappings, and well-defined likelihoods. Although this study considers a limited set of satellites, these properties make them promising for an extension to additional sensors in the GPM constellation.

	\section{Proposed Method}

	\newcommand{\flow}{\Phi}
	
	\subsection{Overview of Conditional Flow Matching}  
	Conditional Flow Matching (CFM)~\cite{flow_matching,  stochastic_interpolant} is a generative modeling approach sharing many similarities with diffusion, and stemming from continuous normalizing flows~\cite{node_nf}. It is a  simulation-free approach, that does not require integrating ordinary differential equations (ODEs) during training, and enables faster sampling than diffusion based approaches. The method involves learning a vector field \(v : [0, 1] \times \Rb^{d} \rightarrow \Rb^{d}\), which defines a time-dependent diffeomorphism, or flow, \(\phi: [0, 1] \times \Rb^{d} \rightarrow \Rb^{d}\), connecting a data space \(\Xc\) to a latent space \(\Zc\). In practice, a neural network \(v_{\theta}(t, \x)\) parameterized by \(\theta\) is trained to approximate the conditional vector field \(u_{t}(\x | \z)\) by minimizing the mean squared error,  
	\begin{equation}
		\Lc_{CFM} = \Eb_{t \sim \Uc({\bf{0}, \bf{I}}), q(\z), p_{t}(\x|\z)} 
		\lVert
		v_{\theta}(t, \x) - u_{t}(\x | \z)
		\rVert_{2}^{2}
		\label{eq:cfm_eq}
	\end{equation}  
	
	\noindent where \(q(\z)\) denotes a Gaussian prior \(\Nc({\bf 0}, {\bf I})\), \(p_{t}(\x|\z)\) is a known probability path that varies for each sample, and $t$ is a random time variable between 0 and 1. Typically, this trajectory interpolates between a data point \(\x_{0}\) and random Gaussian noise \(\z\) as:  
	\begin{equation}
		\x = (1 - t) \x_{0} + t \z.
		\label{eq:ot_displacement_map}
	\end{equation}  
	
	\noindent To generate samples, one first draws a latent code \(\z \sim q(\z)\) and then integrates the ODE using the learned vector field backwards through time to obtain a corresponding data sample $\x$. Conversely, mapping an existing image $\x$ to the latent space involves forward integration of the ODE, producing a latent representation $\z$.  In the remainder of this work so as to avoid confusion, we will refer to this method simply as \emph{flow matching}, while the term \emph{conditional} will specify conditioning on class labels \(\y\), as described in the next section.
	
	\begin{figure*}[!b]
		\centering
		\includegraphics[width=0.84\textwidth]{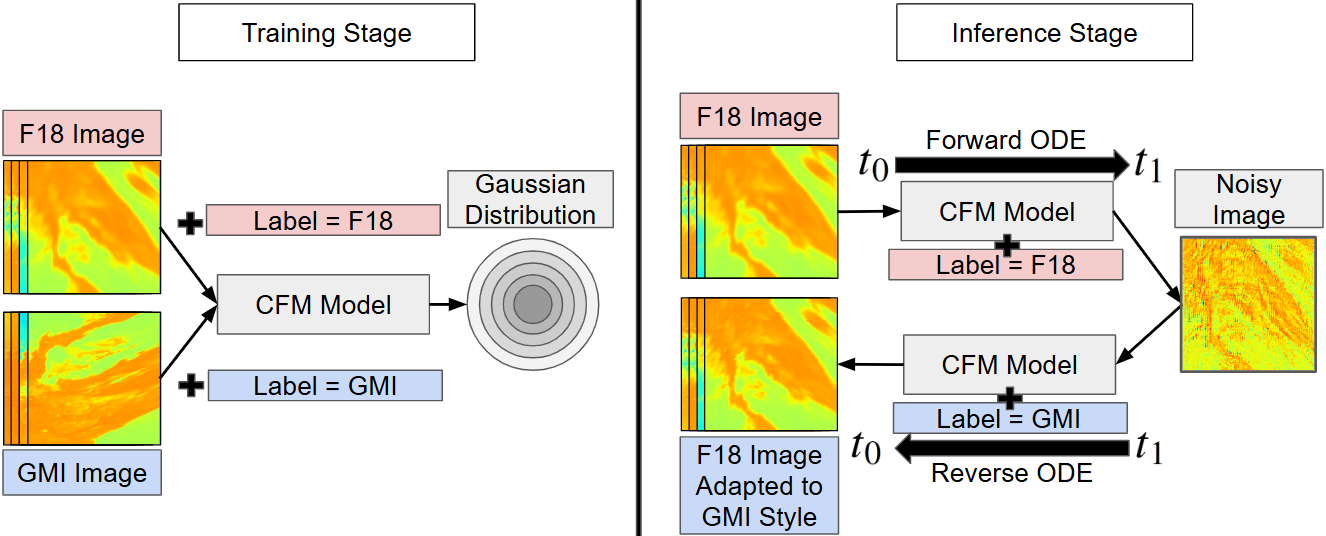}
		\caption{Overview of the proposed conditional flow matching framework. During training (left), the model learns to map both GMI and F18 images to a standard Gaussian distribution using domain-specific labels. During inference (right), domain adaptation is performed in two steps: first, an F18 image is noised by running the forward ODE from $t_{0}$ to $t_{1}$ with the associated label conditioning ($\y_{i}$ in Eq.~\ref{eq:flow_map_uda_used}). Afterwards, the reverse ODE is solved from $t_{1}$ to $t_{0}$ using the GMI conditioning parameters instead ($\y_{j}$ in Eq.~\ref{eq:flow_map_uda_used}), so as to obtain the adapted image in the GMI domain (this corresponds to $\bar{\x}_{i \rightarrow j}$ in Eq.~\ref{eq:flow_map_uda_used}).}
		\label{fig:train_and_inference}
	\end{figure*}

	\subsection{Conditioning and Formalization}
	To jointly model the distribution of multiple satellites within our generative backbone, we adopt a class-conditional formulation, where each satellite is assigned a unique label $\y_i$, with $i \in \{1, 2, \dots, K\}$, and where $K$ denotes the total number of distinct satellites. The generative model can be conditioned to learn satellite-specific data distributions within a unified framework by minimizing the following loss across all image-label pairs $(\x, \y)$
	\begin{equation}
		\Lc_{CFM} = \Eb_{t \sim \Uc({\bf{0}, \bf{I}}), q(\z), p_{t}(\x|\z)}
		\left\lVert
		v_{\theta}(t, \x, \y) - u_{t}(\x \mid \z)
		\right\rVert_{2}^{2}.
		\label{eq:cfm_eq_cond}
	\end{equation}
	Consequently, during sampling, each ODE integration step is conditioned on the satellite class $\y$, allowing the model to generate samples consistent with the specified sensor characteristics. \\
	\indent To formalize conditional sampling and latent-space mapping, we introduce a conditional flow map $\flow_{\theta}(t_{0}, t_{1}, \x, \y)$, which can be interpreted as an ordinary differential equation solver, parameterized by our conditional neural network $v_{\theta}$. This flow transports the image $\x_{t_{0}}$ from an initial time
	$t_{0}$ to a terminal time $t_{1}$, so that
	\begin{equation}
		\x_{t_{1}}=\flow_\theta(t_{0},t_{1},\x_{t_{0}},\y).
		\label{eq:flow_map}
	\end{equation}
	
	\noindent By setting $t_{0}=0$ and $t_{1}=1$, this formulation results in the entire forward ODE mapping an image $\x$ to its latent representation $\z$ so that 
	
	\begin{equation}
		\z =\flow_\theta(t_{0}=0,t_{1}=1,\x,\y).
		\label{eq:flow_map_forward}
	\end{equation}
	Conversely, by setting $t_{0}=1$ and $t_{1}=0$, we get the reverse ODE that maps a latent noise $\z$ to its image $\x$
	\begin{equation}
		\x =\flow_\theta(t_{0}=1,t_{1}=0,\z,\y).
		\label{eq:flow_map_reverse}
	\end{equation}

	\noindent Since the flow mapping is bijective, the original image can be reconstructed exactly by composing the forward and reverse mappings. In particular, we have
	\begin{equation}
		\x = \flow_\theta(t_{0}=1,t_{1}=0, \flow_\theta(t_{0}=0,t_{1}=1,\x,\y),\y),
		\label{eq:flow_map_forward_reverse}
	\end{equation}
	which reconstructs the exact same image if we ignore numerical imprecisions. This property is especially desirable in the context of domain adaptation, as it enables the preservation of the semantic content of an image when transferring between different satellite modalities. \\
	\indent Finally, to perform domain adaptation from any two distinct satellites with labels $\y_{i}$ and  $\y_{j}$ (where $i \neq j$), we modify the conditioning used in the reverse mapping from Eq.~\ref{eq:flow_map_forward_reverse} so that it corresponds to the target label $\y_{j}$. Moreover, we explicitly regulate the intensity of $t_{1}$ (\emph{e.g.} $t_{1}=0.5$, ensuring  $t_{1} < 1$), which controls the intensity of the noising applied during the forward mapping. These design choices lead to the following formulation for the domain adaptation,
	\begin{equation}
		\bar{\x}_{i \rightarrow j} =
		\flow_\theta\!\left(
		t_{1}, t_{0},
		\flow_\theta\!\left(t_{0}, t_{1}, \x, \y_{i}\right),
		\y_{j}
		\right),
		\label{eq:flow_map_uda_used}
	\end{equation}
	where $\x$ denotes the original input image with label $\y_{i}$, and $\bar{\x}_{i \rightarrow j}$  denotes the adapted image which underwent two consecutive flow maps with respective labels  $\y_{i}$ and $\y_{j}$.  In our experiments, $\y_{i}$ and $\y_{j}$ correspond to the F18 and GPM satellites respectively, and $\bar{\x}_{i \rightarrow j}$ corresponds to the F18 image adapted to the GMI domain. Figure~\ref{fig:train_and_inference} summarizes the proposed training and inference stage. Overall, the proposed flow-matching-based formulation enables deterministic and optimal noising and denoising for domain adaptation.
	
	\section{Experiments}
	\noindent \textbf{Datasets.} The training dataset consists of all GMI and F18 observations acquired in 2018. Following \cite{drain_extended, cycle_gan_drain_f18_gpm} only frequencies useful for precipitation estimation are kept. These correspond to 36.6GHz and 89.0GHz for GMI, and 37.0GHz and 91.6GHz for the SSMIS/S instrument. Both vertical and horizontal polarisations are included, resulting in input images with four channels; whose main differences are summarized in Table~\ref{tab:gpm_f18_channels_info}.  To ensure consistency between sensors, polar regions are excluded from the SSMI/S data (latitudes above $70^\circ$N or below $70^\circ$S), so as to match the spatial coverage of GMI. The test dataset contains all overlaps between the GMI and F18 satellite images from the year 2019, that have more than 50 pixels with rain rates higher than 3mm/h. This corresponds to roughly $18 \times 10^{6}$ adapted, collocated pixels from the F18 to the GMI domain.
	\begin{table}[!h]
		\centering
		\resizebox{1.0\linewidth}{!}{%
			\begin{tabular}{lcccc} 
				\toprule
				Instrument                             & Central Frequency & Polarisation & Bandwidth & Pixel Size (km)          \\ 
				\midrule
				\rowcolor[rgb]{0.922,0.961,1} GMI      & 36.6GHz         & V, H         & 1000 MHz  & $6.0 \times 13.4$   \\
				\rowcolor[rgb]{0.922,0.961,1} SSMI/S & 37.0GHz          & V, H         & 1580 MHz  & $25.0 \times 12.5$  \\ 
				\midrule
				\rowcolor[rgb]{1,0.95,0.95} GMI      & 89.0GHz          & V, H         & 6000 MHz  & $3.0 \times 13.4$   \\
				\rowcolor[rgb]{1,0.95,0.95} SSMI/S & 91.6GHz        & V, H         & 2829 MHz  & $12.5 \times 12.5$  \\
				\bottomrule
			\end{tabular}
		}
		\caption{Technical information on the considered GMI and SSMI/S, grouped by frequency bands and polarisations (V vertical, H horizontal).}
		\label{tab:gpm_f18_channels_info}
	\end{table}

	\noindent \textbf{Evaluation Metrics.} To assess the performance of our method, we employ the DRAIN framework~\cite{drain_extended} to estimate raining rates from the adapted F18-to-GMI images. Those are then compared with RRs obtained from the original GMI images. To ensure pixels between these images are matched, a KDTree algorithm is applied to find the nearest neighbors within a region of 10km, before calculating distance metrics such as the mean error (ME), mean absolute error (MAE), and the root mean squared error (RMSE). Besides precipitation estimation, we also perform a qualitative assessment of the adapted images. \\
	\noindent \textbf{Implementation Details.} \textcolor{commentcolor}{Experiments are conducted using a class-conditional DiT \cite{dit}, trained for 50 epochs using the Adam optimizer with a learning rate of $1.0 \times 10^{-4}$ and a batch size of 64. The model uses patch embeddings of size 4, 12 attention heads across 12 transformer layers, and a hidden dimension of 384, resulting in a total of 32.91M parameters.} During training, input images are randomly cropped to a resolution of $128 \times 128$ and augmented with random horizontal and vertical flips. In practice, when performing domain adaptation, we set the time variables $t_{0}$ and $t_{1}$ from Eq.~\ref{eq:flow_map_uda_used}, to respectively 0.0 and 0.5. For the ODE solver, we use the default Dormand-Prince method from the torchdiffeq library~\cite{node_nf}.\\ 
    \textcolor{commentcolor}{\noindent \textbf{Baselines and Comparative Methods.}
    For the baseline, we use the non-adapted F18 images as direct input to our generative model.} \textcolor{commentcolor}{\noindent For the comparative methods, we evaluate against three approaches: (i) \textit{stochastic diffusion noising} (as in \cite{sdedit}), where Eq.~\ref{eq:ot_displacement_map} is applied with random Gaussian noise to perturb samples at $t=0.5$; (ii) \textit{full ODE noising} (as in \cite{ddib}), which is a special case of our formulation with $t_{1}=1.0$ in Eq.~\ref{eq:flow_map_uda_used}; and (iii) \textit{CycleGAN} (from \cite{cycle_gan_drain_f18_gpm}), which also focused on adapting F18 images to the GMI domain.}

\newcommand{\subfigwbis}{0.30}

\begin{table}[!b]
\centering
\renewcommand{\arraystretch}{0.90}
\resizebox{\linewidth}{!}{%
\begin{tabular}{llccc} 
\toprule
RR in mm/h   & Model                                   & RMSE            & MAE             & ME                \\ 
\hline
$[0.0, 0.1)$ & F18 Baseline                            & 0.074           & 0.011           & 0.011             \\
             & SDEdit~\cite{sdedit}                    & 0.147           & 0.019           & 0.019             \\
             & DDIB~\cite{ddib}                        & 0.102           & 0.012           & 0.012             \\
             & CycleGAN~\cite{cycle_gan_drain_f18_gpm} & 0.101           & 0.016           & 0.016             \\
             & Ours                                    & \textbf{0.071}  & \textbf{0.007}  & \textbf{0.007}    \\ 
\hline
$[0.1, 2.5)$ & F18 Baseline                            & \textbf{0.638}  & \textbf{0.412}  & -0.267            \\
             & SDEdit~\cite{sdedit}                    & 0.873           & 0.520           & -0.279            \\
             & DDIB~\cite{ddib}                        & 0.852           & 0.510           & -0.277            \\
             & CycleGAN~\cite{cycle_gan_drain_f18_gpm} & 0.697           & 0.423           & -0.234            \\
             & Ours                                    & 0.706           & 0.427           & \textbf{-0.210}   \\ 
\hline
$[2.5, 10)$  & F18 Baseline                            & 3.652           & 3.201           & -3.016            \\
             & SDEdit~\cite{sdedit}                    & 4.050           & 3.479           & -2.935            \\
             & DDIB~\cite{ddib}                        & 4.007           & 3.426           & -2.840            \\
             & CycleGAN~\cite{cycle_gan_drain_f18_gpm} & 3.817           & 3.333           & -3.116            \\
             & Ours                                    & \textbf{3.479}  & \textbf{2.867}  & \textbf{-2.342}   \\ 
\hline
$[10, 50)$   & F18 Baseline                            & 17.480          & 15.545          & -15.353           \\
             & SDEdit~\cite{sdedit}                    & 17.388          & 15.401          & -15.122           \\
             & DDIB~\cite{ddib}                        & 17.050          & 14.968          & -14.009           \\
             & CycleGAN~\cite{cycle_gan_drain_f18_gpm} & 17.316          & 15.366          & -14.981           \\
             & Ours                                    & \textbf{15.836} & \textbf{13.736} & \textbf{-12.851}  \\ 
\hline
RR $\geq 50$ & F18 Baseline                            & 58.807          & 57.951          & -57.951           \\
             & SDEdit~\cite{sdedit}                    & 58.997          & 58.139          & -58.139           \\
             & DDIB~\cite{ddib}                        & 55.579          & 54.073          & -53.150           \\
             & CycleGAN~\cite{cycle_gan_drain_f18_gpm} & 57.703          & 56.529          & -56.529           \\
             & Ours                                    & \textbf{53.581} & \textbf{51.929} & \textbf{-51.494}  \\ 
\hline\hline
RR $\geq 0$  & F18 Baseline                            & 0.609           & 0.086           & -0.052            \\
             & SDEdit~\cite{sdedit}                    & 0.665           & 0.104           & -0.045            \\
             & DDIB~\cite{ddib}                        & 0.640           & 0.095           & -0.049            \\
             & CycleGAN~\cite{cycle_gan_drain_f18_gpm} & 0.619           & 0.092           & -0.045            \\
             & Ours                                    & \textbf{0.575}  & \textbf{0.079}  & \textbf{-0.043}   \\
\hline
\end{tabular}
}
\caption{RMSE, MAE, and ME for different rain rate intervals.}
\label{tab:rain_metrics}
\end{table}
\begin{figure*}[!h]
    \centering
    % First row: 3 subfigures
    \begin{subfigure}{\subfigwbis\textwidth}
        \centering
        \includegraphics[width=\linewidth]{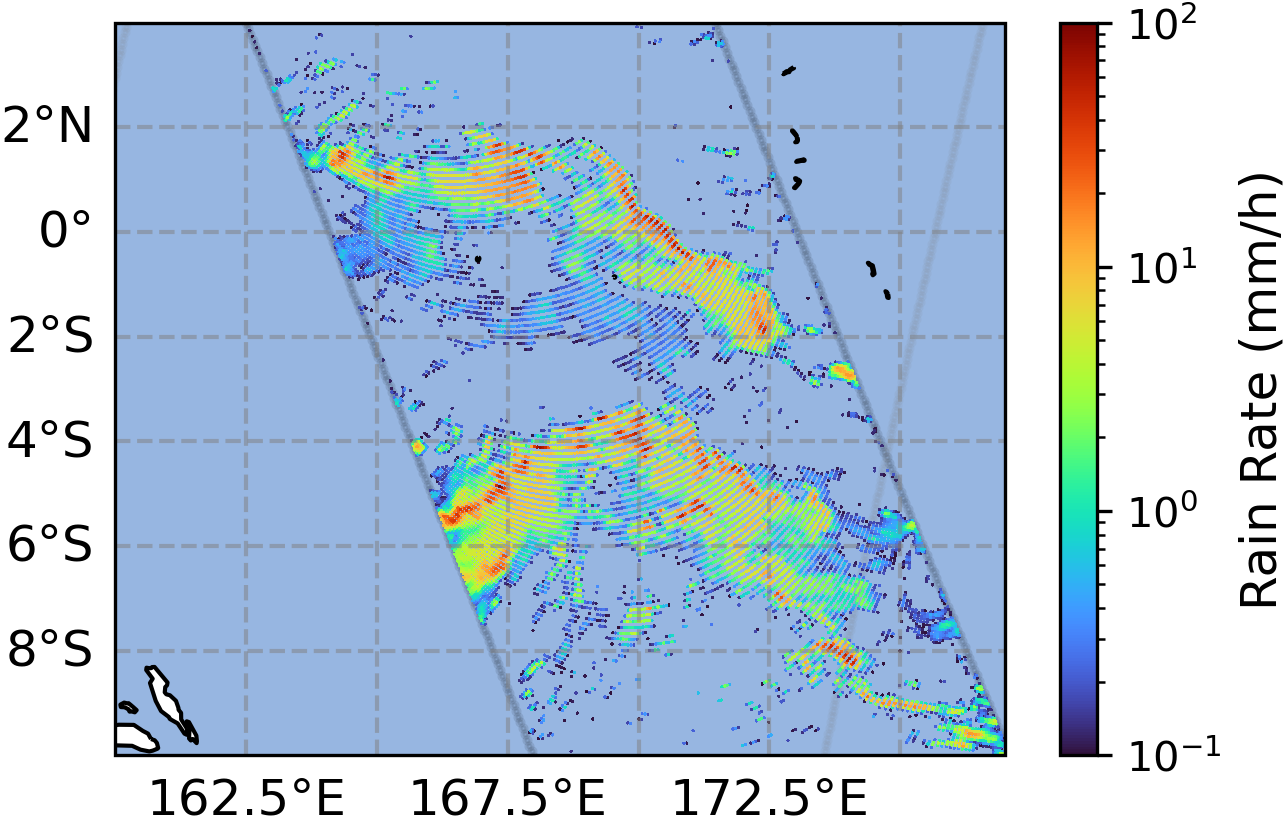}
        \caption{DRAIN on GMI (Ground-truth)}
        \label{fig:rrr_gpm}
    \end{subfigure}
    \hfill
    \begin{subfigure}{\subfigwbis\textwidth}
        \centering
        \includegraphics[width=\linewidth]{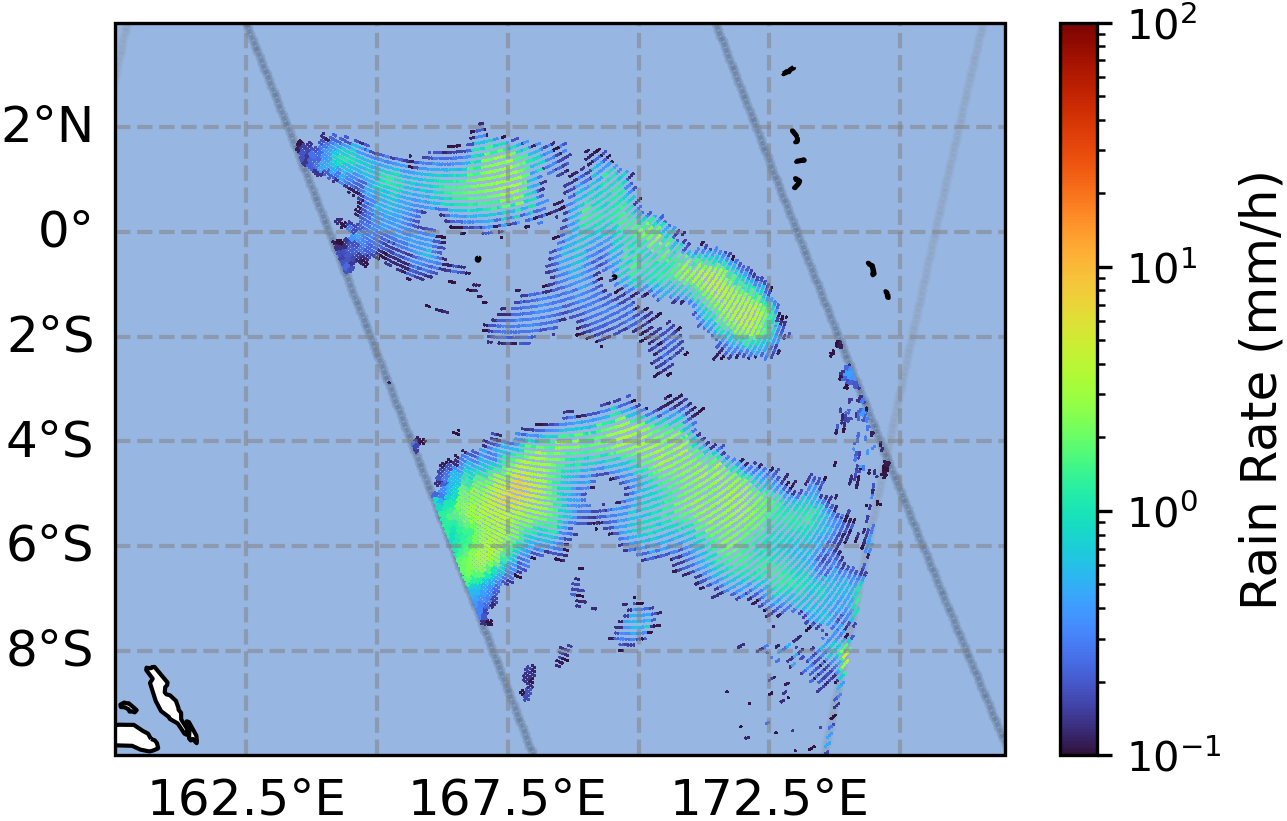}
        \caption{DRAIN on F18 (Baseline)}
        \label{fig:rrr_cfm}
    \end{subfigure}
    \hfill
    \begin{subfigure}{\subfigwbis\textwidth}
        \centering
        \includegraphics[width=\linewidth]{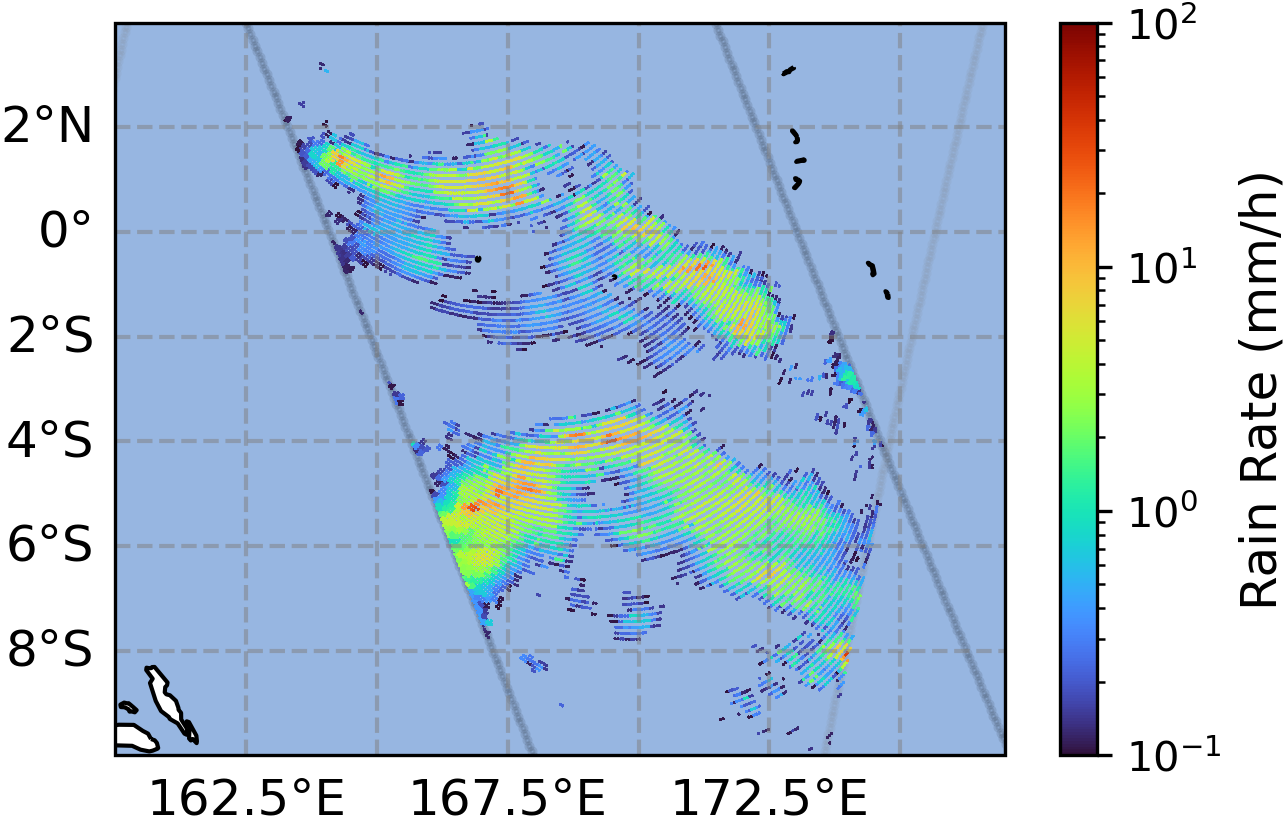}
        \caption{DRAIN on Adapted F18 (Ours)}
        \label{fig:rrr_gpm_dup}
    \end{subfigure}
    % Second row: 3 subfigures
    \begin{subfigure}{\subfigwbis\textwidth}
        \centering
        \includegraphics[width=\linewidth]{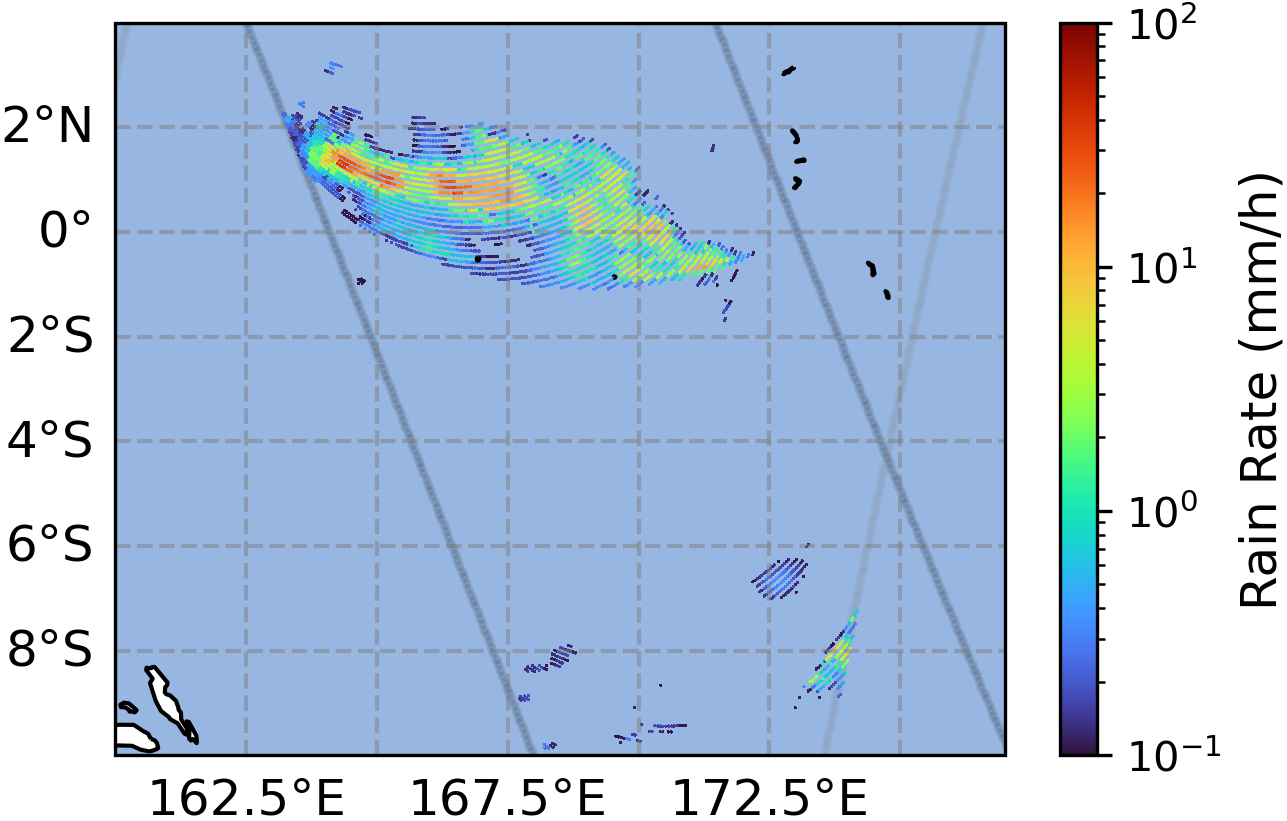}
        \caption{DRAIN on Adapted F18 \\ (DDIB~\cite{ddib})}
        \label{fig:rrr_cyclegan}
    \end{subfigure}
    \hfill
    \begin{subfigure}{\subfigwbis\textwidth}
        \centering
        \includegraphics[width=\linewidth]{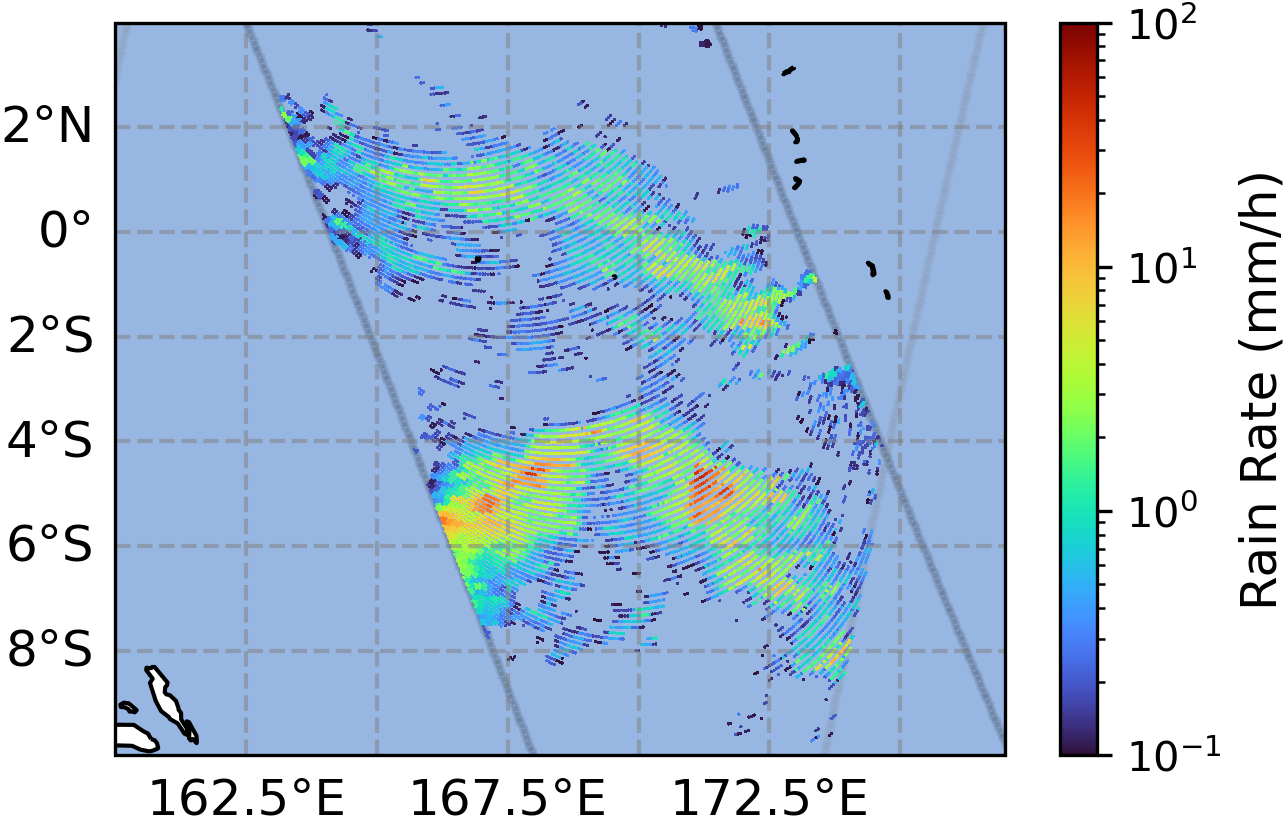}
        \caption{DRAIN on Adapted F18 \\ (SDEdit~\cite{sdedit})}
        \label{fig:rrr_f18}
    \end{subfigure}
    \hfill
    \begin{subfigure}{\subfigwbis\textwidth}
        \centering
        \includegraphics[width=\linewidth]{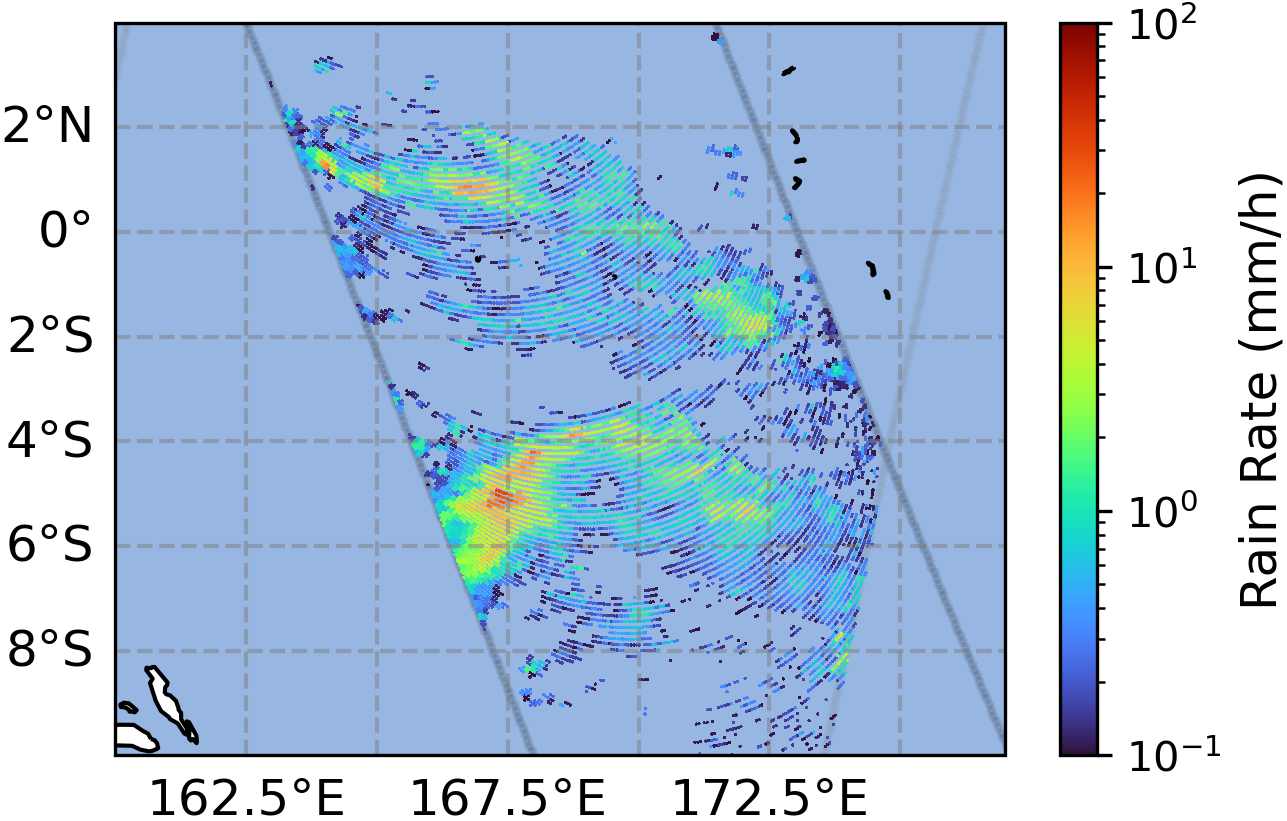}
        \caption{DRAIN on Adapted F18 \\ (CycleGAN~\cite{cycle_gan_drain_f18_gpm})}
        \label{fig:rrr_cfm_dup}
    \end{subfigure}
    \caption{\textcolor{commentcolor}{Comparison of precipitation estimates (mm/h) obtained using DRAIN on the ground-truth GMI (\subref{fig:rrr_gpm}), the unadapted F18 baseline (\subref{fig:rrr_cfm}), our method (\subref{fig:rrr_gpm_dup}), and the comparative methods DDIB~\cite{ddib}, SDEdit~\cite{sdedit}, and CycleGAN~\cite{cycle_gan_drain_f18_gpm} (\subref{fig:rrr_cyclegan}, \subref{fig:rrr_f18}, \subref{fig:rrr_cfm_dup}). Results are shown for the 2019-02-06 overpass all projected on GMI pixels. Note that perfect collocation is unattainable due to differences in observation angles, acquisition times, and pixel counts, which may cause discrepancies despite \emph{``perfect''} adaptation.}}

    \label{fig:rrr_comparison}
\end{figure*}

\newcommand{\subfigw}{0.18}

\begin{figure*}[!h]
    \centering
    \begin{subfigure}[b]{\subfigw \textwidth}
        \centering
        \includegraphics[width=\textwidth]{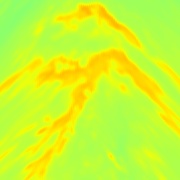}
        \caption[]{F18 \\ Baseline}
        \label{fig:tb_f18}
    \end{subfigure}
    \hfill
    \begin{subfigure}[b]{\subfigw \textwidth}
        \centering
        \includegraphics[width=\textwidth]{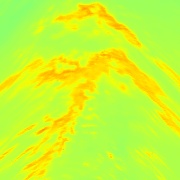}
        \caption[]{F18 to GMI \\ (Ours)}
        \label{fig:tb_ours}
    \end{subfigure}
    \hfill
    \begin{subfigure}[b]{\subfigw \textwidth}
        \centering
        \includegraphics[width=\textwidth]{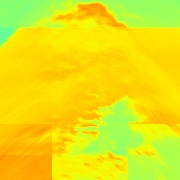}
        \caption[]{F18 to GMI \\ (DDIB~\cite{ddib})}
        \label{fig:tb_ddib}
    \end{subfigure}
    \hfill
    \begin{subfigure}[b]{\subfigw \textwidth}
        \centering
        \includegraphics[width=\textwidth]{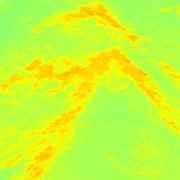}
        \caption[]{F18 to GMI \\ (SDEdit~\cite{sdedit})}
        \label{fig:tb_sdedit}
    \end{subfigure}
    \hfill
    \begin{subfigure}[b]{\subfigw \textwidth}
        \centering
        \includegraphics[width=\textwidth]{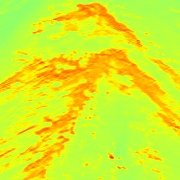}
        \caption[]{F18 to GMI \\ (CycleGAN~\cite{cycle_gan_drain_f18_gpm})}
        \label{fig:tb_gan}
    \end{subfigure}
    \caption{\textcolor{commentcolor}{Comparison between the original F18 images (37GHz-V channel) (\subref{fig:tb_f18}), with the adapted images generated using our proposed method (\subref{fig:tb_ours}), as well as the comparative methods DDIB~\cite{ddib} (\subref{fig:tb_ddib}), SDEdit~\cite{sdedit} (\subref{fig:tb_sdedit}), and CycleGAN~\cite{cycle_gan_drain_f18_gpm} (\subref{fig:tb_gan}).}}
    \label{fig:tb_comparison}
\end{figure*}
	
	\subsection{Performance Comparison.}
	Table~\ref{tab:rain_metrics} shows the performance of the proposed unsupervised domain adaptation method, when compared to the F18 baseline, \textcolor{commentcolor}{and the comparative methods}.  Overall the proposed method achieves highest performance on moderate and heavy rainfalls. This is valuable because the heavy-tailed rain distribution is the least present in satellite images, meaning the proposed model generalized very well.  This is confirmed in Figure~\ref{fig:rrr_comparison}, where we compare precipitation estimates on the adapted F18-to-GMI images, with the ground truth GMI, as well as with \textcolor{commentcolor}{the comparative methods and the unadapted F18 images}. Finally, in Figure~\ref{fig:tb_comparison}, we directly show the radiometer images before and after being adapted. It can be observed that the resolution improves significantly compared to the original F18 images, revealing much finer details. Additionally, our proposed method is less prone to hallucinating details that were not initially present in the original images, unlike competing methods.

	\section{Conclusion}
	In this paper, we presented a novel approach for unsupervised domain adaptation using conditional flow matching. The strength of our method lies in its ability to perform precise, content-aware modifications through invertible and gradual transformations governed by conditional ODEs. Our experiments demonstrate the benefits of this unsupervised domain adaptation approach, highlighting its effectiveness in two-domain scenarios — and showcasing great promise for multiple domains. As future work, we aim to extend this method to handle multiple satellite instruments both within and beyond the GPM-Core constellation, further exploring its potential for large-scale multi-domain adaptation.

	\bibliographystyle{IEEEbib}
	% If using any of the following journal options:
	%   wet, dap, dce, eds, prm, flw, jdm, psy, rsm
	% then use the \printbibliography line instead of:
	\bibliography{_refs/satellite, _refs/deep_learning}
	%\printbibliography

\end{document}